\documentclass{article}

\usepackage[square,numbers]{natbib}
\usepackage[main, final]{neurips_2026}
\usepackage{graphicx}
\usepackage{refcount}
\usepackage{color}
\usepackage{makecell}
\usepackage{multirow}
\usepackage{amsmath}
\usepackage[utf8]{inputenc} 
\usepackage[T1]{fontenc}    
\usepackage{hyperref}       
\usepackage{url}            
\usepackage{booktabs}       
\usepackage{amsfonts}       
\usepackage{nicefrac}       
\usepackage{microtype}      
\usepackage{xcolor}         
\usepackage{soul, color}
\usepackage{amsmath}
\usepackage{graphicx}
\usepackage{enumitem}
\usepackage{multirow}
\usepackage{subcaption}
\usepackage{adjustbox}

\title{PIU-CR: Physics-Informed Deep Unfolding with SAR-Optical Image Fusion for Cloud Removal}

\author{%
	Gengque Fan$^1$,\, Han Xu$^1$\thanks{Corresponding author},  \, Junxi Li$^2$,\, Hao Zhang$^3$,\, Guangcan Liu$^1$,\, Jiayi Ma$^4$\\
	$^1$ School of Automation, Southeast University, Nanjing, China\\
	$^2$ School of Instrument Science and Engineering, Southeast University, Nanjing, China\\
	$^3$ Electronic Information School, Wuhan University, Wuhan, China\\
	$^4$ School of Robotics, Wuhan University, Wuhan, China\\
	\texttt{\{gengquefan, xu\_han, lijunxi\}@seu.edu.cn, zhpersonalbox@gmail.com},\\ \texttt{guangcanliu@seu.edu.cn, jyma2010@gmail.com}
}

\begin{document}

\maketitle

\begin{abstract}
  Multi-modal cloud removal exploits complementary optical and synthetic aperture radar (SAR) images to restore cloud-degraded remote sensing images. Most existing methods rely on black-box, data-driven architectures with limited interpretability of cross-modal interactions. Although some attempts introduce physical priors into networks, their oversimplified formulations fail to faithfully characterize the underlying physical imaging process and the networks remain loosely coupled with optimization variables. To address these issues, we propose a physics-informed deep unfolding network for cloud removal via optical-SAR image fusion, termed as PIU-CR. We first formulate cloud removal as a physics-informed optimization problem. Specifically, we fundamentally ground the optimization in the physical imaging mechanism of cloud degradation by embedding the atmospheric scattering model into the data fidelity term. Then, the regularization terms establish an explicit cross-modal interaction mechanism, overcoming the single-modal information scarcity and the obscure, uninterpretable interactions in black-box models. They are dedicated to SAR-guided structure and spectral regularization, jointly introducing cross-modal structural feature alignment and coordinated spectral consistency constraints. The resulting optimization problem is then unfolded into a multi-stage neural network with dedicated structure and spectral branches. Each stage corresponds to an explicit optimization iteration step, enabling interpretable and effective reconstruction process. Experiments demonstrate that the proposed method outperforms state-of-the-art methods. The code is publicly available at \href{https://github.com/Fgq188/PIU\_CR}{https://github.com/Fgq188/PIU\_CR}.
\end{abstract}

\section{Introduction}
During image acquisition in remote sensing, cloud formation, as a prevalent atmospheric phenomenon, frequently disrupts optical imaging process. By scattering and absorbing the reflected radiation from Earth's surface, inherent and pervasive cloud coverage severely degrades or even completely occludes ground information. It compromises the quality and reliability of optical remote sensing data. Hence, enabling robust cloud removal to restore cloud-free, high quality optical images is indispensable in practical remote sensing. It critically enables downstream applications such as land-cover classification~\cite{use4,use5,use7,innovation}, environmental monitoring~\cite{use1,use6,use8,use9}, and change detection~\cite{use2,use3}.


A straightforward category is to reconstruct the cloud-free image solely from a single-modal cloudy optical image. Early single-modal methods rely on hand-crafted priors~\cite{earlyprior1,earlyprior2,earlyprior3} or physical models~\cite{earlymodel1,earlymodel2,earlymodel3} to reconstruct the cloud-covered underlying image. More recently, deep learning-based methods~\cite{hsformer,dens2cr,cloudruler,earlymodel3,single2,single3,single4} have significantly improved performance by leveraging powerful feature extraction and nonlinear modeling capabilities. However, regardless of method design, a single optical image inherently suffers from missing information under severe cloud coverage. The corruption or elimination of the signal renders reliable reconstruction impossible and hinders accurate recovery of structural details and spectral fidelity~\cite{gcepanet}.


\begin{figure}
	\centering
	\includegraphics[width=1\linewidth]{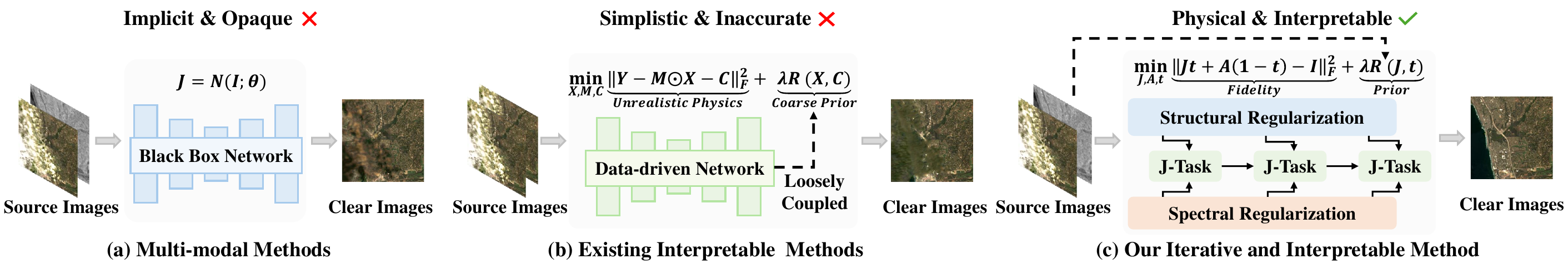}\\
	\caption{Illustration of remote sensing cloud removal methods.}
	\label{introduction}
\end{figure}


Recent methods explore cloud removal via multi-modal data fusion. An ideal auxiliary modality should be insensitive to clouds and atmospheric conditions, providing reliable structural information where optical signals are lost. Synthetic aperture radar (SAR), with cloud-penetrating capability, has been widely adopted. In fusion, optical images preserve spectral characteristics, while SAR supplies robust structural cues. Existing multi-modal methods follow an end-to-end learning paradigm: SAR and optical images are jointly fed into a neural network (ranging from CNN~\cite{dens2cr,glfcr,hs2p}, Transformer~\cite{hpn,ussrn}, Mamba~\cite{mamba1,mamba2}, to Diffusion~\cite{diffcr,emrdm} architectures) to directly reconstruct the cloud-free image. Regardless of architecture design, the fusion process is implicitly learned without explicit physical modeling, remaining data-driven and mathematically intractable. It limits interpretability and hinders analysis of cross-modal interactions and guidance in the final reconstruction.

To address the interpretability of deep learning frameworks, several works attempt to incorporate physical priors, as in Fig.~\ref{introduction}(b). They typically formulate cloud removal as a simplified decomposition problem under low-rank or sparse priors on cloud-free and cloud components~\cite{lrrssn,stlr}. On this basis, neural networks aim to learn implicit image priors or parameterize parts of optimization process. \textit{From the perspective of physical modeling}, these methods simplify cloud degradation as additive clean image and cloud components. Although optimization-friendly, it does not reflect the actual imaging process of clouds as the degradation is governed by atmospheric scattering, where scene radiance and atmospheric effects are jointly involved. The simplified modeling fails to capture the realistic physical relationship under complex cloud conditions. Meanwhile, the regularization strategies (\textit{e.g.}, low-rank or sparsity priors) rely on coarse statistical assumptions, oversimplifying data structures and inadequately describing complex spatial and spectral variations in remote sensing imagery. \textit{From the perspective of model design}, although neural networks are introduced to enhance representation capacity, the overall framework still relies on data-driven parameter fitting. The introduced neural networks are not explicitly aligned with the optimization variables or physical components in the degradation model. As a result, the role of each network module in the reconstruction process remains unclear, leading to limited interpretability. Furthermore, the interaction between the optimization procedure and neural networks is loosely defined, where the network mainly serves as an auxiliary regularizer rather than being inherently embedded into the optimization process.

To overcome these limitations, we propose a novel cloud removal method based on multi-modal data fusion. Specifically, we formulate the problem as an optimization framework under a physics-informed model and further unfold it into a neural network with cross-modal SAR–optical guidance. Firstly, compared with simplified formulations, cloudy image formation is a more physically realistic process. Scene radiance is attenuated by the atmosphere while atmospheric light is added through a transmission-dependent scattering process. This process can be mathematically expressed by the atmospheric scattering model~\cite{asm}. Therefore, to better align with the physical imaging process, we incorporate this model into the data fidelity term in our framework, which leading to more consistent and reliable reconstruction of cloud-free images. Secondly, SAR image can provide robust structural cues under cloud occlusion and is insensitive to atmospheric interference, making it a reliable complementary source for scene recovery. Based on this property, we introduce SAR-guided structure and spectral regularization terms into the optimization framework. This enables effective exploitation of cross-modal priors for cloud-covered regions. Finally, we unfold the optimization process into a multi-stage neural network, where each stage corresponds to one iteration step. The underlying physical mechanisms and multi-modal constraints are explicitly unfolded into corresponding iterative network modules. This design ensures that the entire reconstruction trajectory is mathematically traceable, thereby preserving interpretability as well as benefiting from the representation ability of neural networks. Contributions can be concluded as:

\begin{itemize}[leftmargin=*]
	\item We propose a novel SAR-guided deep unfolding network for cloud removal, which enables effective cross-modal guidance from structure to spectral reconstruction in a physically interpretable manner. To the best of our knowledge, it is the first work to explicitly incorporate SAR image into a deep unfolding framework for this task.
	\item We formulate the data fidelity term based on the atmospheric scattering model, which explicitly characterizes the imaging formation process of cloudy observations. This physically grounded formulation provides an interpretable constraint consistent with the cloud imaging process.
	\item We design both structure and spectral regularization terms based on SAR image to further enhance reconstruction quality. The optimization process is further unfolded into a multi-stage neural network, thereby combining the advantages of model-based and data-driven methods.
\end{itemize}

\section{Related Works}
\label{related_works}

\paragraph{Cloud Removal Methods.}
Learning-based cloud removal methods can be categorized as single-modal and multi-modal methods. The former mainly utilize optical imagery, which can be further divided into single-image methods and multi-temporal methods. Single-image methods learn an end-to-end mapping from cloudy observations to cloud-free images~\cite{hsformer,dens2cr,cloudruler,single2,single3}. Due to the single observation, they suffer from insufficient information in heavily cloud-covered regions. Multi-temporal methods exploit multiple optical images acquired at different times over the same region~\cite{diffcr,time1,time2}. By leveraging temporal redundancy, they can partially recover cloud-covered areas. Nevertheless, their performance depends heavily on cloud-free observations, and they may introduce inconsistencies or artifacts when significant scene changes occur. To address this limitation, multi-modal methods leverage complementary data from different sensors. SAR-based methods have shown strong potential due to cloud-penetrating capability, enabling more reliable reconstruction under heavy cloud coverage~\cite{hpn,emrdm}. However, these data-driven methods generally lack explicit physical constraints and theoretical interpretability, limiting robustness and reliability. Some methods attempt to combine deep learning with physical interpretability~\cite{lrrssn,stlr} . However, their formulations are often oversimplified, failing to accurately model complex cloud formation heavy cloud conditions.


\paragraph{Deep Unfolding Network in Computer Vision.}
Deep unfolding networks~\cite{deepundolfing1,deepunfolding2,deepunfolding3,iterrf} bridge the gap between model-based optimization and data-driven learning by unrolling iterative algorithms into trainable network architectures, offering both interpretability and theoretical grounding. Owing to these advantages, they have been widely applied in various computer vision tasks, including low-light enhancement, pan-sharpening and so on. For instance, URetinex-Net~\cite{uretinex} introduces a Retinex-inspired deep unfolding framework that transforms the iterative optimization process of low-light enhancement into a learnable network. DISPNet~\cite{pan} introduces a deep unfolding network integrated with intrinsic supervision for pan‑sharpening task to produce high-quality images. Despite these advancements, deep unfolding methods still require further exploration in remote sensing cloud removal tasks, especially for multi-modal scenarios.

\section{Method}
\label{method}

\subsection{Problem Formulation}
Modeling cloudy observations based on physical mechanisms is desirable for describing cloud formation and degradation. Consistent with the physical imaging process, the atmospheric scattering model represents the observed optical image as a combination of attenuated scene radiance and scattered atmospheric light. Based on this model, the cloudy image can be mathematically formulated as~\cite{asm1,asm2,asm3,asm4}:
\begin{equation}
	I=J\cdot t+A\cdot (1-t),
\end{equation}
where $I$ is the observed cloudy optical multi-spectral (MS) image. $J$ is the cloud-free image to be recovered. $A$ is the global atmospheric light, which is a constant for one image. $t$ is the transmission map describing the portion of scene radiance that reaches sensor without scattering, with values ranging from 0 to 1. $J\cdot t$ models the direct attenuation of scene radiance. $A\cdot (1-t)$ is the airlight caused by atmospheric scattering.

However, recovering $J$, $A$, and $t$ from a single optical image $I$ is inherently ill-posed, as the solution is non-unique. To obtain a stable and physically meaningful solution, additional prior constraints based on physical properties are required to regularize the estimation of these variables. Thus,  the estimation can be formulated as the following inverse minimization problem:
\begin{equation}
\label{eq_0}
    \arg\min_{J,A,t} \| J t + A(1-t) - I \|_F^2 + \lambda_A R(A) + \lambda_t R(t) + \lambda_J R(J),
\end{equation}
where the first term is the data fidelity term, with $\|\cdot \|_F^2$ denoting the Frobenius norm. $R(A)$, $R(t)$ and $R(J)$ denote regularization terms imposed on $A$, $t$ and $J$. $\lambda_A$, $\lambda_t$, $\lambda_J$ control the trade-off.

For the atmospheric light $A$, it is the global atmospheric light and spatially invariant for each image. Thus, it can be reliably estimated from the data fidelity term without the need to introduce constraints $R(A)$. Thus, the regularization terms are mainly imposed on $t$ and $J$ to constrain the solution space. For the transmission map $t$, due to the continuous distribution of atmospheric effects, $t$ should vary smoothly in space. Therefore, a smooth constraint $R(t) = \| \nabla t \|_F^2$ is imposed for its stabilization.


For the cloud-free image $J$, solely relying on a single optical observation remains insufficient. When the cloud is thick, the transmission $t$ approaches zero and the observed image is dominated by atmospheric light, leading to severe information loss in $J$. As mentioned above, the SAR image $S$ exhibits cloud-penetrating and is insensitive to atmospheric conditions, providing reliable structural information under cloud occlusion. Moreover, regions with similar structural patterns in $S$ are likely to correspond to similar land-cover types, further implying consistent spectral characteristics. Therefore, we incorporate $S$ as a physical prior guidance, then SAR and optical images can be jointly exploited to improve reconstruction identifiability.

Thus, we adopt the SAR image $S$ into the regularization term of latent image $J$ to constrain the overall optimization. Under the guidance of $S$, $R(J)$ is extended to $R(J,S)$ to explicitly encode cross-modal relationships into optimization and enable physically consistent reconstruction in cloud-covered regions. Thus, the objective problem in Eq.~\eqref{eq_0} is revised as:
\begin{equation}
\arg\min_{J,A,t} \| J t + A(1-t) - I \|_F^2 + \lambda_t R(t) + \lambda_J R(J,S).\label{revise_obj}
\end{equation}
At this point, the remaining core issue lies in the definite construction of regularization term $R(J,S)$.

Since cloud removal aims to recover a cloud-free image that is not only structurally faithful to the underlying scene but also spectrally consistent with real land-cover characteristics, the reconstruction process should be constrained in both spatial and spectral domains. As discussed above, SAR imagery can provide reliable structural guidance under cloud occlusion, while similar structural patterns further imply potential spectral consistency in the optical image across regions. Motivated by this observation, we construct $R(J,S)$ from both structure-aware and spectral-aware perspectives, enabling SAR information to guide the reconstruction process in a physically meaningful manner.

\textbf{Structural Alignment Regularization.} A common assumption in cross-modal processing is that features from different modalities of the same scene are aligned to be consistent at structural level~\cite{pan}. Since SAR and cloudy optical images depict the same scene, the structural patterns in SAR $S$ are expected to correlate with those of cloud-free image $J$. Thus, we impose a structure alignment constraint as a regularization term to preserve structural fidelity during reconstruction as:
\begin{equation}
	R_{struc}(J, S) = \| T_S S - T_J J \|_F^2,
\end{equation}
where $T_S$ and $T_J$ represent the structure feature extraction networks for SAR and MS images respectively, which are implemented as learnable neural networks.

\begin{figure}[!t]
	\centering
	\includegraphics[width=1\linewidth]{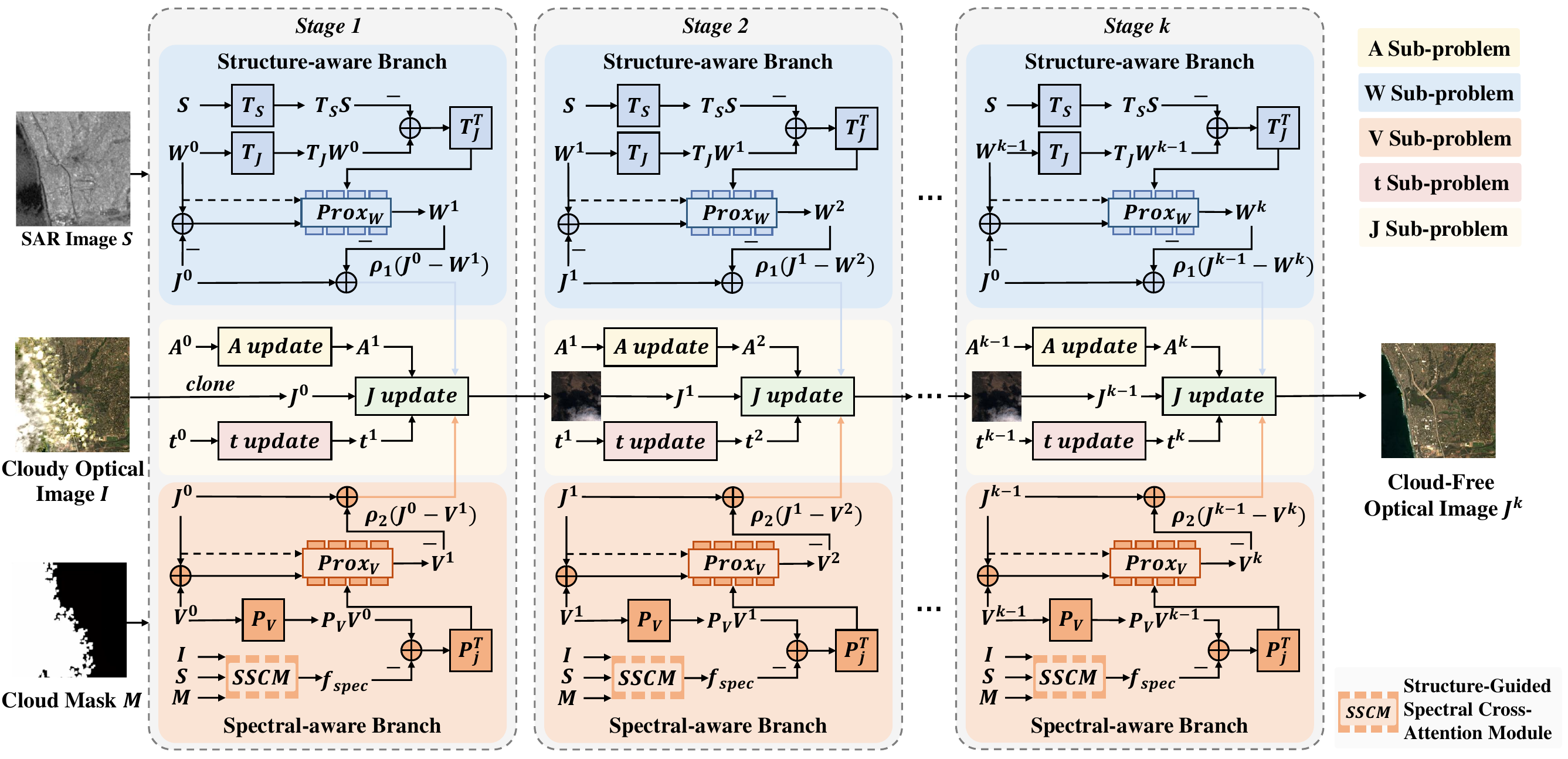}\\
	\caption{Overall framework of the proposed physics-informed deep unfolding cloud removal method with SAR-optical image fusion (PIU-CR).}
	\label{model}
\end{figure}

\textbf{Spectral Alignment Regularization}. In MS images, regions with similar structural characteristics are likely to correspond to the same type of land-cover, \textit{i.e.}, tend to exhibit similar spectral responses. If a cloud-covered region shares similar structural patterns with a cloud-free region, they are assumed to represent the same underlying land-cover, thus exhibiting consistent spectral properties. Meanwhile, SAR image can provide structural information of the entire scene. To exploit this property, we propose a SAR-guided spectral alignment regularization that focuses on cloud-covered regions. Specifically, a cloud mask $M$ with ones representing cloudy pixels is first employed to identify cloud-covered regions. Then, the structural similarity is computed between cloud-covered and cloud-free regions through structural features $T_S S$. Subsequently, we impose spectral constraint on the regions with high structural similarity. Thus, a spectral consistency constraint is imposed during optimization:
\begin{equation}
	R_{spec}(J, S) = \| \omega M \odot ( \Phi(T_S S, M, P_I I) - (P_J J) ) \|_F^2,
\end{equation}
where $\omega$ is the confidence weighting factor. $\Phi$ is a structure-guided spectral cross-attention module. $P_I,P_J$ are spectral feature extraction networks for $I$ and $J$, respectively.

With the incorporation of these priors, Eq.~\eqref{revise_obj} can be reformulated as follows, which corresponds to the proposed model structure shown in Fig.~\ref{model}, with each term explicitly mapped to its corresponding module in the network:
\begin{align}
	\arg\min_{J,A,t} &\| J t + A(1-t) - I \|_F^2
	+ \lambda_t \| \nabla t \|_F^2
	+ \lambda_{struc} \| T_S S - T_J J \|_F^2
	\nonumber
	\\& + \lambda_{spec} \| \omega M \odot ( \Phi(T_S S, M, P_I I) - (P_J J) ) \|_F^2.
	\label{eq_min}
\end{align}

\subsection{Iterative Unfolding Optimization}
Due to the strong coupling of $J$ with other variables in the optimization problem, it is difficult to directly solve Eq.~\eqref{eq_min}. We adopt half-quadratic splitting (HQS)~\cite{HQS}, decoupling the original problem into a set of tractable subproblems by introducing auxiliary variables. Specifically, two auxiliary variables $W,V$ are introduced to separately handle structure and spectral regularization terms as:
\begin{align}
\label{eq_all}
	\arg\min_{J,A,t,W,V} \| J t + A(1-t) - I \|_F^2
	+ \lambda_t \| \nabla t \|_F^2
	+ \lambda_{struc} \| T_S S - T_J W \|_F^2
	\nonumber
	\\ + \lambda_{spec} \| \omega M \odot ( \Phi(T_S S, M, P_I I) - (P_J V) ) \|_F^2 + \rho_1 \| W - J \|_F^2 + \rho_2 \| V - J \|_F^2,
\end{align}
where $\rho_1$ and $\rho_2$ are penalty parameters. To solve Eq.~\eqref{eq_all}, we adopt an alternating optimization strategy. Each variable is updated sequentially while keeping others fixed. It results in a series of subproblems corresponding to $A$, $W$, $V$, $t$, and $J$, which are iteratively refined at each step $k$.

We begin by updating the atmospheric light $A$, as it only appears in the data fidelity term and admits a simple quadratic form. The corresponding subproblem is given by:
\begin{equation}
\label{eqA}
A^{k} = \arg\min_{A} \| J^{k-1} t^{k-1} + A(1-t^{k-1}) - I \|_F^2,
\end{equation}
which yields a closed-form solution $A^k = \frac{\sum_x \big(1-t^{k-1}\big)\big(I-J^{k-1}t^{k-1}\big)}{\sum_x \big(1-t^{k-1}\big)^2+\epsilon}$, where $x$ is the spatial location of each pixel, and the summation is performed over all spatial pixels. Moreover, to improve numerical stability when $t$ approaches 1, we introduce a small constant $\epsilon$ into the denominator in our implementation.

After updating $A$, we proceed to optimize the auxiliary variable $W$, which is introduced to approximate the structure-constrained reconstruction. The corresponding subproblem is given by:
\begin{equation}
	W^{k} = \arg\min_{W} \lambda_{struc} \| T_S S - T_J W \|_F^2 + \rho_1 \| W - J^{k-1} \|_F^2.
\end{equation}
Similarly, the auxiliary variable $V$ is updated to incorporate the SAR-guided spectral constraint:
\begin{equation}
	V^{k} = \arg\min_{V} \lambda_{spec} \| \omega M \odot( \Phi(T_S S, M, P_I I) - (P_J V)) \|_F^2 + \rho_2 \| V - J^{k-1} \|_F^2.
\end{equation}

Both $W$- and $V$-subproblems involve composite objectives with implicit regularization operators. Thus, we adopt the proximal gradient projection method to iteratively solve them while preserving the corresponding regularization properties. The update rules are formulated as:
\begin{equation}
	W^{k} = prox_W( W^{k-1} - \delta_1( 2\lambda_{struc} T_J^T(T_J W^{k-1} - T_S S) + 2\rho_1(W^{k-1} - J^{k-1}) ) ),
\end{equation}
\begin{equation}
	\footnotesize V^{k} = prox_V( V^{k-1} - \delta_2( 2\lambda_{spec} P_J^T \omega^2 M^2 (P_J V^{k-1} - \Phi(T_S S, M, P_I I)) + 2\rho_2(V^{k-1} - J^{k-1}) ) ),
\end{equation}
where $prox_W$ and $prox_V$ denote the proximal operators. $\delta_1$ and $\delta_2$ are step sizes.

With the auxiliary variables updated, we then optimize the transmission map $t$, which controls the attenuation effect in the atmospheric scattering model. This subproblem is formulated as:
\begin{equation}
	t^{k} = \arg\min_{t} \| J^{k-1} t + A^{k}(1-t) - I \|_F^2 + \lambda_t \| \nabla t \|_F^2,
\end{equation}
Finally, $J$ is updated under joint guidance of data fidelity term and reconstructed auxiliary variables:
\begin{equation}
	J^{k} = \arg\min_{J} \| J t^{k} + A^{k}(1-t^{k}) - I \|_F^2 + \rho_1 \| W^{k} - J \|_F^2 + \rho_2 \| V^{k} - J \|_F^2.
\end{equation}
Since the objectives of the $t$-subproblem and $J$-subproblem are smooth and differentiable, we adopt gradient descent to progressively refine them. Their update rules are given by:
\begin{equation}
	t^{k} = t^{k-1} - \eta_1 (2(J - A)(Jt^{k-1} + A(1-t^{k-1}) - I) + 2 \lambda_t \nabla^T \nabla t^{k-1}),
\end{equation}
\begin{equation}
	J^{k} = J^{k-1} - \eta_2 (2t( J^{k-1} t + A(1-t) - I ) + 2\rho_1(J^{k-1} - W) + 2\rho_2(J^{k-1} - V)),
\end{equation}
where $\eta_1$ and $\eta_2$ denote the corresponding step sizes. Moreover, to map $t$ into a valid transmission range from 0 to 1, a sigmoid activation function is applied during the update.

Through this alternating update process, all variables are progressively refined, and the overall optimization is effectively approximated via a sequence of interpretable iterative steps.

\subsection{Network Architectures}

The proposed deep unfolding network is composed of $k$ stages, each corresponding to one iteration of the optimization procedure. The overall architecture comprises two major components: the structure-aware branch and the spectral-aware branch.

\textbf{Structure-aware Branch.} The structure-aware branch corresponds to the structure alignment regularization term in the optimization formulation, which aims to enforce spatial consistency between the reconstructed image and the observed input. Specifically, $T_S$ and $T_J$are implemented as convolutional feature extract networks, while $T_J^T$ is realized as a corresponding inverse mapping network. The proximal operator $prox_W$ is modeled as a lightweight projection network. To avoid introducing excessive parameters and preserve the interpretability of the optimization process, each module adopts a compact design with three $3 \times 3$ convolutional layers and nonlinear activations.This simple design ensures sufficient representation capacity while improving training stability and maintaining consistency with the operator formulation in the deep unfolding framework.

\textbf{Spectral-aware Branch.} The spectral-aware branch is derived from the spectral regularization guided by cross-modal information. Similar to the structure-aware branch, the involved operators are parameterized as learnable modules. Specifically, $P_I$, $P_J$ and $P_J^T$ are implemented as convolutional mapping networks and their corresponding inverse mapping network, while the proximal operator $prox_V$ is realized as a lightweight projection network. In addition, to enforce the spectral regularization in cloud-covered regions, we introduce a structure-guided spectral cross-attention module (SSCM), denoted as $\Phi$ (see more in \textit{Supplementary Material}). This module leverages structural features to identify similarity between cloud-covered and cloud-free regions and provide spectral constraint via a cross-attention mechanism, where the query, key, and value are defined as:
%
%
\begin{equation}
	Q = T_s S, \quad K = T_s S \odot (1-M), \quad V = P_I I \odot (1-M).
\end{equation}
The output structure-guided spectral feature maps are $f_{spec} = \mathrm{softmax}\left( \frac{QK^T}{\sqrt{d}} \right) V$, where $d$ is the feature dimension. Since the spectral information in cloud-free regions is already reliable, the constraint is applied only to cloud-covered areas by incorporating the cloud mask $M$, which selectively enforces consistency on reconstructed spectral features. Furthermore, to capture the reliability of structural matching, a confidence weight $\omega$ is computed from query-key similarity.

\subsection{Loss Functions}
To constrain both spatial and spectral properties of the reconstructed image, we define a joint loss function  consisting of $L_1$, structural similarity, and spectral angle mapper losses. With $\alpha$, $\beta$ and $\gamma$ controlling the trade-off, it is defined as:
\begin{equation}
	\mathcal{L} = \alpha \mathcal{L}_{L_1} + \beta \mathcal{L}_{SSIM} + \gamma \mathcal{L}_{SAM}.
\end{equation}
$L_1$ loss preserves overall intensity by enforcing pixel-wise fidelity and avoid over-smoothing as:
\begin{equation}
	\mathcal{L}_{L_1} = \| J^{k} - GT \|_1,
\end{equation}
where $J^{k}$ is the output of the $k^{th}$ stage. $GT$ is the reference optical image.
\newline
To maintain spatial structures and textures, SSIM loss measures structural similarity as:
\begin{equation}
	\mathcal{L}_{SSIM} = 1 - SSIM(J^{k}, GT).
\end{equation}
To maintain spectral similarity and preserve the spectral angle relationships in the reconstructed image, SAM loss enforces spectral consistency by measuring the angular difference between spectral vectors. With $i$ denoting the $i^{th}$ pixel, it is defined as:
\begin{equation}
	\mathcal{L}_{SAM} = \frac{1}{N} \sum_{i=1}^{N} \arccos (
	\frac{ \langle J_i^k, GT_{i} \rangle }
	{ \|J_i^k\|_2 \cdot \|GT_{i}\|_2 }).
\end{equation}

\section{Experiments}
\label{others}

\textbf{Datasets and Implementation}. We conduct experiments on SEN12MS-CR~\cite{sen12mscr} and SMILE-CR~\cite{smile} datasets. SEN12MS-CR contains 122,218 patch triplets with 13-band Sentinel-2 cloudy/cloud-free optical images and dual-polarization Sentinel-1 SAR data, covering diverse scenes and four seasons. We use 20,000 samples for training and 7,899 for testing. SMILE-CR consists of 6-band Landsat-8 cloudy/cloud-free optical images with corresponding SAR data. Regarding the acquisition of cloud masks, for the SEN12MS-CR dataset, we follow the official preprocessing pipeline released with the dataset to generate cloud masks. For the SMILE-CR dataset, the cloud masks are directly provided by the dataset. More details are provided in \textit{Supplementary Material}.

During training, images are resized to a resolution of 256 $\times$ 256. The number of iteration stages $K$ is 3. Hyperparameters $\alpha$, $\beta$, and $\gamma$ are 1, 1, and 0.2, respectively. Gradient descent step sizes $\eta_1$ and $\eta_2$ are 0.1 and 0.05, respectively. $\rho_1$ and $\rho_2$ are initialized to 1 and progressively increased with the increase of iteration number. All experiments are conducted on an NVIDIA RTX 4090 GPU with a batch size of 16, using the Adam optimizer. The training epochs for SEN12MS-CR and SMILE-CR datasets are set to 150 and 900, respectively.


\textbf{Comparison Results}. We compare PIU-CR with several state-of-the-art (SOTA) cloud removal methods, including single-modal cloud removal methods (AMGAN~\cite{amgan}, ACANet~\cite{acanet}, HSFormer~\cite{hsformer}, and CloudRuler~\cite{cloudruler}) and multi-modal cloud removal methods (DSen2-CR~\cite{dens2cr}, GLF-CR~\cite{glfcr}, HS2P~\cite{hs2p}, USSRN~\cite{ussrn}, HPN-CR~\cite{hpn}, and EMRDM~\cite{emrdm}). In our experiments, cloud masks are provided only to the methods that are originally designed to use them (i.e., DSen2-CR and AMGAN), while other methods follow their original input configurations to ensure a fair comparison. Moreover, the cloud mask only identifies relatively reliable cloud-free regions to provide spatial guidance for spectral regularization in our framework, rather than directly determining the restoration results.

\begin{figure*}[!t]
	\centering
	\includegraphics[width=\linewidth]{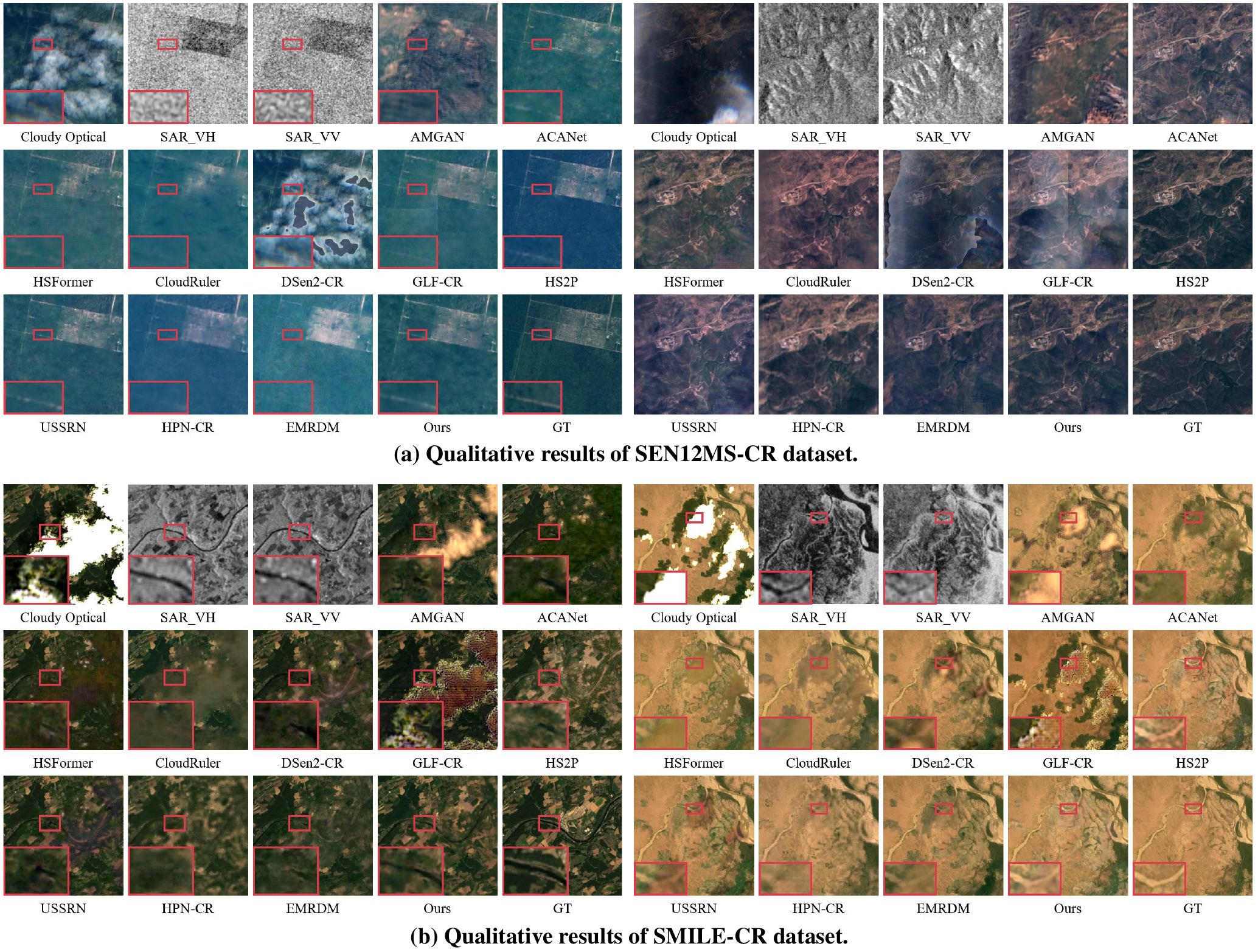}\\
  \vspace{-0.05in}
	\caption{Qualitative comparison of different methods on \textbf{SEN12MS-CR} and \textbf{SMILE-CR} datasets.}
	\label{comparision_results}
\end{figure*}

\textit{Qualitative Results.} Results on two datasets are shown in Fig.~\ref{comparision_results}. First, benefiting from the structure and spectral regularization terms, the proposed method achieves clear advantages in both structural detail preservation and spectral reconstruction. As shown in the first group of Fig.~\ref{comparision_results}(a), our method better restores the structural textures of underlying objects, while the second group demonstrates its superiority in spectral recovery. Second, compared with single-modal methods, multi-modal approaches generally show improved performance, especially in heavily cloud-occluded regions. As shown in the first group of Fig.~\ref{comparision_results}(b), single-modal methods often struggle to recover the underlying objects due to insufficient information, whereas our method effectively leverages SAR guidance to reconstruct cloud-covered areas. Finally, different from existing multi-modal methods that implicitly fuse SAR information, our method explicitly incorporates SAR image into the optimization framework as regularization terms, enabling more effective and controllable utilization of SAR-derived structural information during reconstruction. More results are shown in \textit{Supplementary Material}.

\begin{table*}[t]
	\caption{Quantitative comparison of different methods on \textbf{SEN12MS-CR} and \textbf{SMILE-CR} datasets.}
  \vspace{-0.05in}
	\centering
	\resizebox{\linewidth}{!}{
    \renewcommand\arraystretch{1.15}
		\begin{tabular}{c@{\,~~}|c@{\,~~~~}|c@{\,~~~~}c@{\,~~~~}c@{\,~~~~}c@{\,~~~~}c@{\,~~~}|
        c@{\,~~~~}c@{\,~~~~}c@{\,~~~~}c@{\,~~~~}c@{\,~~~}}
			\toprule
			\multirow{2}{*}{\textbf{Types}} & \textbf{\textbf{Datasets}} & \multicolumn{5}{c|}{\textbf{SEN12MS-CR}} & \multicolumn{5}{c}{\textbf{SMILE-CR}} \\
			\cmidrule(lr){2-7} \cmidrule(lr){8-12}
			& \textbf{Methods} & PSNR$\uparrow$ & SSIM$\uparrow$ & MAE$\downarrow$ & RMSE$\downarrow$ & SAM$\downarrow$
			& PSNR$\uparrow$ & SSIM$\uparrow$ & MAE$\downarrow$ & RMSE$\downarrow$ & SAM$\downarrow$ \\
			\midrule
			\multirow{4}{*}{\textbf{Single-modal}}
			& AMGAN~\cite{amgan}      & 23.621 & 0.801 & 0.040 & 0.072 & 0.226 & 32.742 & 0.855 & 0.029 & 0.042 & 0.098 \\
			& ACANet~\cite{acanet}     & 22.061 & 0.845 & 0.039 & 0.089 & 0.254 & 34.253 & 0.898 & 0.023 & 0.034 & 0.066 \\
			& HSFormer~\cite{hsformer}   & 27.934 & 0.873 & 0.029 & 0.043 & 0.164 & \underline{42.071} & 0.904 & 0.016 & 0.025 & 0.063 \\
			& CloudRuler~\cite{cloudruler} & 27.256 & 0.877 & 0.031 & 0.050 & 0.182 & 37.028 & 0.905 & 0.017 & 0.025 & 0.069 \\
			\midrule
			\multirow{6}{*}{\textbf{Multi-modal}}
			& DSen2-CR~\cite{dens2cr}   & 20.709 & 0.788 & 0.047 & 0.102 & 0.291 & 40.568 & 0.903 & 0.016 & 0.027 & 0.059 \\
			& GLF-CR~\cite{glfcr}     & 26.795 & 0.869 & 0.029 & 0.049 & 0.156 & 29.326 & 0.805 & 0.044 & 0.063 & 0.149 \\
			& HS2P~\cite{hs2p}       & 27.952 & 0.878 & 0.031 & 0.044 & 0.157 & 40.149 & \underline{0.919} & \underline{0.014} & \underline{0.022} & \underline{0.054} \\
			& USSRN~\cite{ussrn}      & 28.412 & \underline{0.883} & 0.029 & 0.042 & 0.154 & 41.408 & 0.915 & 0.015 & 0.024 & 0.056 \\
			& HPN-CR~\cite{hpn}     & 28.242 & 0.869 & 0.030 & 0.042 & \underline{0.152} & 32.463 & 0.874 & 0.023 & 0.031 & 0.088 \\
			& EMRDM~\cite{emrdm}      & \underline{28.692} & 0.873 & \underline{0.029} & \underline{0.041} & 0.153 & 37.277 & 0.910 & 0.018 & 0.026 & 0.063 \\
			\cmidrule(lr){1-12}
			\multirow{1}{*}{\textbf{Multi-modal}}
			& \textbf{Ours}
			& \textbf{28.856} & \textbf{0.889} & \textbf{0.027} & \textbf{0.039} & \textbf{0.144}
			& \textbf{42.713} & \textbf{0.929} & \textbf{0.013} & \textbf{0.019} & \textbf{0.050} \\
			\bottomrule
		\end{tabular}%
	}
	\label{model_performance}
\end{table*}

\textit{Quantitative Results.} Evaluation is conducted with peak signal-to-noise ratio (PSNR), SSIM, mean absolute error (MAE), root mean square error (RMSE), and SAM. As shown in Tab.~\ref{model_performance}, our method achieves the best performance across all metrics on both datasets. Highest SSIM values indicate that our method preserves spatial structures and fine details. SAM focuses on spectral consistency, where lower values imply that the reconstructed spectral are closer to ground truth. The remaining metrics measure pixel-wise reconstruction accuracy, indicating more precise intensity recovery. The overall improvements across these metrics are attributed to the integration of physically grounded modeling and multi-modal information. On the one hand, the atmospheric scattering model enforces physical consistency. On the other hand, the explicit incorporation of SAR image in the optimization framework enables effective utilization of cross-modal interaction, leading to better performance.

\begin{figure*}[!t]
	\centering
	\includegraphics[width=\linewidth]{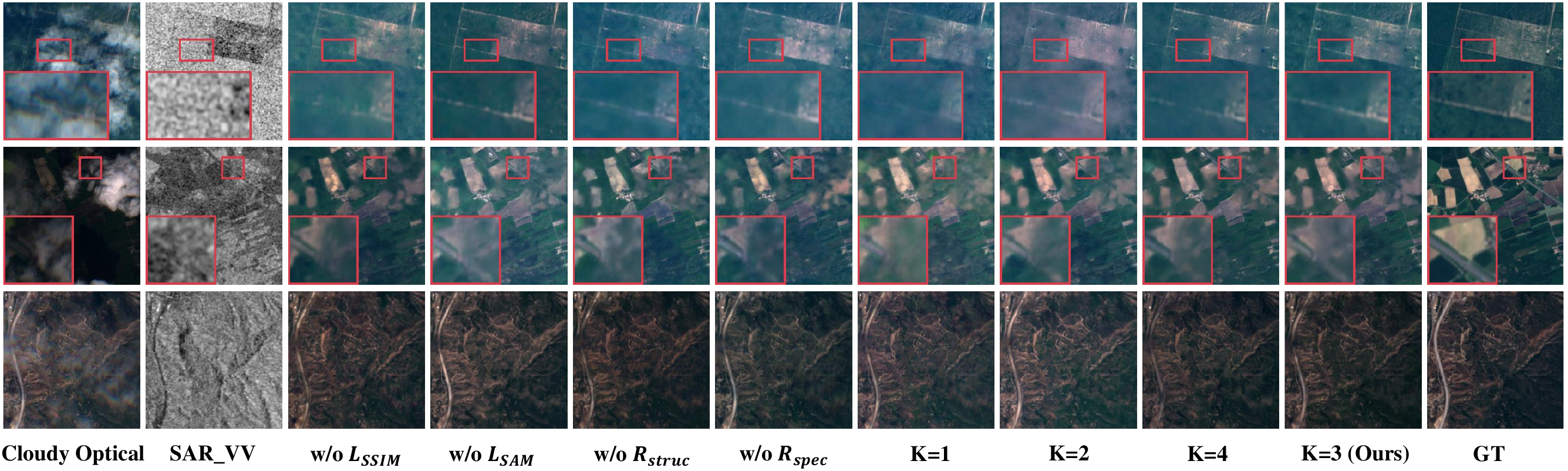}\\
\vspace{-0.06in}
	\caption{Qualitative results of ablation study.}
	\label{ablation_visual}
\end{figure*}

\textbf{Ablation Study}. We conducted extensive ablation studies on the loss functions, optimization architecture, number of iterations, and hyperparameter selection, as shown in Fig.~\ref{ablation_visual} and Tab.~\ref{ablation_metrics}.

\textit{Losses and Regularization Terms.} As shown in the third to sixth columns of Fig.~\ref{ablation_visual} and Tab.~\ref{ablation_metrics}(a), removing SSIM loss or structure regularization term leads to blurred structural details in cloud-covered regions. Removing SAM loss or spectral regularization term results in noticeable color distortion. It demonstrates that both structural and spectral constraints are crucial in preserving spatial details and spectral consistency.

\textit{Iteration Numbers.} According to the iteration number, we evaluate different numbers of iteration stages, as shown in the seventh to ninth columns of Fig.~\ref{ablation_visual} and Tab.~\ref{ablation_metrics}(b). The reconstruction quality gradually improves as the number of iterations increases. However, increasing the number of iterations to four does not yield further improvements over three iterations, while introducing additional computational overhead. Therefore, we set $K$=3 in our final model, achieving the best reconstruction performance while maintaining computational efficiency.

\textit{Hyperparameters.} Furthermore, we conduct hyperparameter ablation experiments, as shown in Tab.~\ref{ablation_metrics}(c), to investigate the effects of $\eta_1$ and $\eta_2$. The combination of $\eta_1=0.1$ and $\eta_2=0.05$ achieves the best overall performance, demonstrating the effectiveness of the selected hyperparameter configuration.

\textit{Spectral Regularization Term.} Moreover, we conducted further studies on spectral regularization, as shown in Tab.~\ref{ablation_metrics}(d). Specifically, replacing the proposed spectral constraint $R_{spec}$ with global spectral feature consistency constraint $\left\| P_I I - P_J J \right\|_F^2$ and replacing SSCM with standard cross-attention consistency both lead to less targeted spectral guidance, highlighting the importance of region-aware spectral interactions between cloud-free and cloud-covered regions.

\begin{table*}[!t]
	\centering
	\captionsetup[subtable]{font=scriptsize}
	\caption{Ablation studies on different components of the proposed method.}
	\label{ablation_metrics}
\vspace{-0.05in}
	\begin{subtable}{0.47\linewidth}
		\centering
\caption{Ablation on different losses and regularization terms.}\vspace{-0.05in}
		\resizebox{\linewidth}{!}{
			\renewcommand\arraystretch{1.15}
			\begin{tabular}{l@{\,~~}|c@{\,~~}c@{\,~~}c@{\,~~}c@{\,~~}c@{\,~}}
				\toprule
				\textbf{Metrics} & \textbf{PSNR} $\uparrow$ & \textbf{SSIM} $\uparrow$ & \textbf{MAE} $\downarrow$ & \textbf{RMSE} $\downarrow$ & \textbf{SAM} $\downarrow$ \\
				\midrule
				w/o $L_{SSIM}$ & 28.151 & 0.877 & 0.029 & 0.042 & 0.154 \\
				w/o $L_{SAM}$  & 28.602 & 0.889 & 0.028 & 0.041 & 0.148 \\
				w/o $R_{struc}$ & 28.481 & 0.883 & 0.029 & 0.041 & 0.150 \\
				w/o $R_{spec}$ & 28.622 & 0.885 & 0.028 & 0.040 & 0.149 \\
				Ours           & \textbf{28.856} & \textbf{0.889} & \textbf{0.027} & \textbf{0.039} & \textbf{0.144} \\
				\bottomrule
		\end{tabular}}
	\end{subtable}
	\hfill
	\begin{subtable}{0.51\linewidth}
		\centering
\caption{Ablation on different iteration numbers.}
\vspace{-0.05in}
		\resizebox{\linewidth}{!}{
			\renewcommand\arraystretch{1.4}
			\begin{tabular}{l@{\,~~}|c@{\,~~}c@{\,~~}c@{\,~~}c@{\,~~}c@{\,~}}
				\toprule
				\textbf{Metrics} & \textbf{PSNR} $\uparrow$ & \textbf{SSIM} $\uparrow$ & \textbf{MAE} $\downarrow$ & \textbf{RMSE} $\downarrow$ & \textbf{SAM} $\downarrow$ \\
				\midrule
				Iteration K=1      & 28.392 & 0.881 & 0.029 & 0.042 & 0.153 \\
				Iteration K=2      & 28.241 & 0.883 & 0.029 & 0.043 & 0.155 \\
				Iteration K=3 (ours) & \textbf{28.856} & \textbf{0.889} & \textbf{0.027} & \textbf{0.039} & \textbf{0.144} \\
				Iteration K=4      & 28.593 & 0.886 & 0.028 & 0.040 & 0.148 \\
				\bottomrule
		\end{tabular}}
	\end{subtable}
	
	\vspace{0.05in}
	\begin{subtable}{0.47\linewidth}
		\centering
\caption{Ablation on hyperparameters $\eta_1$ and $\eta_2$.}\vspace{-0.05in}
		\resizebox{\linewidth}{!}{
			\renewcommand\arraystretch{1.15}
			\begin{tabular}{l@{\,~~}|c@{\,~~}c@{\,~~}|c@{\,~~}c@{\,~~}c@{\,~~}c@{\,~~}c@{\,~}}
				\toprule
				\textbf{Config} & $\eta_1$ & $\eta_2$ & \textbf{PSNR} $\uparrow$ & \textbf{SSIM} $\uparrow$ & \textbf{MAE} $\downarrow$ & \textbf{RMSE} $\downarrow$ & \textbf{SAM} $\downarrow$ \\
				\midrule
				I     & 0.1  & 0.01 & 28.741 & 0.887 & 0.028 & 0.041 & 0.148 \\
				II    & 0.1  & 0.1  & 28.793 & 0.888 & 0.028 & \textbf{0.039} & 0.145 \\
				III   & 0.05 & 0.05 & 28.810 & 0.886 & 0.028 & 0.040 & 0.148 \\
				IV    & 0.2  & 0.05 & 28.722 & 0.887 & 0.028 & 0.040 & 0.146 \\
				Ours  & 0.1  & 0.05 & \textbf{28.856} & \textbf{0.889} & \textbf{0.027} & \textbf{0.039} & \textbf{0.144} \\
				\bottomrule
		\end{tabular}}
	\end{subtable}
	\hfill
	\begin{subtable}{0.51\linewidth}
		\centering
\caption{Ablation on spectral regularization term $R_{spec}$.}
\vspace{-0.05in}
		\resizebox{\linewidth}{!}{
			\renewcommand\arraystretch{1.2}
			\begin{tabular}{l@{\,~~}|c@{\,~~}c@{\,~~}c@{\,~~}c@{\,~~}c@{\,~}}
				\toprule
				\textbf{Metrics} & \textbf{PSNR} $\uparrow$ & \textbf{SSIM} $\uparrow$ & \textbf{MAE} $\downarrow$ & \textbf{RMSE} $\downarrow$ & \textbf{SAM} $\downarrow$ \\
				\midrule
				w/o $R_{spec}$ & 28.622 & 0.885 & 0.028 & 0.040 & 0.149 \\
				Alternative $R_{spec}$ & 28.795 & 0.886 & 0.028 & 0.040 & 0.148 \\
				Alternative SSCM & 28.830 & 0.887 & 0.028 & 0.040 & 0.147 \\
				Ours & \textbf{28.856} & \textbf{0.889} & \textbf{0.027} & \textbf{0.039} & \textbf{0.144} \\
				\bottomrule
		\end{tabular}}
	\end{subtable}
\end{table*}

\begin{figure*}[!t]
	\centering
	\includegraphics[width=\linewidth]{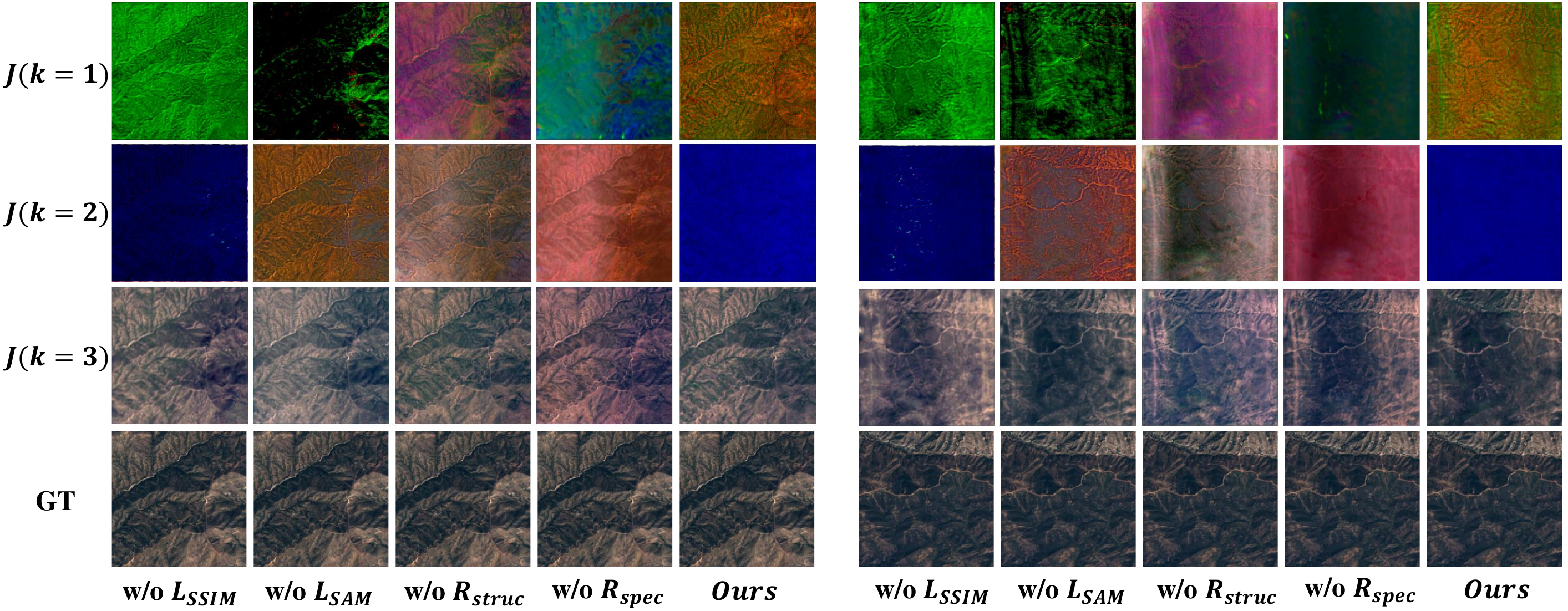}\\
	\caption{Visualization of optimization variable $J$ across different iterations.}
	\label{iter_analysis}
\end{figure*}

\textbf{Interpretability Analysis.} Fig.~\ref{iter_analysis} visualizes two reconstructed results of the optimization variable $J$ across different iterations, as well as the results with certain modules removed. It can be observed that, as the iteration proceeds, the quality of $J$ is progressively improved. The cloud-covered regions gradually evolve from blurred and distorted appearances to clearer structures with more accurate spectral information, demonstrating the effectiveness of the iterative optimization process in deep unfolding. In contrast, removing key modules leads to noticeable degradation, including loss of structural details and increase of spectral distortion. Overall, these results highlight that the proposed method achieves a physics-guided and interpretable optimization process, where each stage corresponds to a meaningful update, enhancing both transparency and reliability.

\textbf{Convergence and Efficiency Analysis.} We conduct convergence analysis for the proposed method in the \textit{Supplementary Material} to demonstrate its good convergence performance. The efficiency is evaluated by parameter numbers and runtime, conducted on a single image with a resolution of 256 $\times$ 256 pixels. The reported runtime represents the complete inference cost of each model under the same input resolution. As reported in Tab.~\ref{params_runtime}, the proposed method achieves superior performance while attaining the shortest runtime and a relatively low (second-best) parameter count compared with the competitors, demonstrating an effective balance between efficiency and performance.

\begin{table*}[!t]
	\caption{Parameters and runtime comparison of different methods.}
   \vspace{-0.05in}
	\centering
	\resizebox{\linewidth}{!}{
		\renewcommand\arraystretch{1.2}
		\begin{tabular}{l@{\,~~}|c@{\,~~}c@{\,~~}c@{\,~~}c@{\,~~}c@{\,~~}c@{\,~~}
c@{\,~~}c@{\,~~}c@{\,~~}c@{\,~~}c@{\,~}}
			\toprule
			\textbf{Methods} & \textbf{AMGAN} & \textbf{ACANet} & \textbf{HSFormer} & \textbf{CloudRuler} & \textbf{DSen2-CR} & \textbf{GLF-CR} & \textbf{HS2P} & \textbf{USSRN} & \textbf{HPN-CR} & \textbf{EMRDM} & \textbf{Ours} \\
			\midrule
			\textbf{Para (M)} & \textbf{0.238} & 20.444 & 5.655 & 6.805 & 37.830 & 14.773 & 21.411 & 18.794 & 3.551 & 39.134 & \underline{1.557} \\
			\textbf{Runtime (s)} & \underline{0.031} & 0.348 & 0.128 & 0.055 & 3.589 & 0.281 & 0.887 & 0.166 & 0.079 & 0.368 & \textbf{0.016} \\
			\bottomrule
		\end{tabular}%
	}
	\label{params_runtime}
\end{table*}

\section{Conclusion}
This paper proposes a physics-informed deep unfolding framework for SAR–optical cloud removal. Based on the atmospheric scattering model, the method formulates a physically consistent and interpretable optimization problem. SAR-guided structural and spectral regularization is introduced to exploit complementary multi-modal information for cloud removal. The optimization is further unfolded into a multi-stage network, combining interpretability with representation ability.


\begin{ack}
This work was supported by the National Natural Science Foundation of China (NSFC) under Grant 62502085 and U21B2027, the Natural Science Foundation of Jiangsu Province of China under Grant BK20241280, and the Start-up Research Fund of Southeast University under Grant RF1028624061 and RF1028623006.
\end{ack}

\bibliography{neurips_2026}

\newpage
\appendix

\section*{Technical Appendices and Supplementary Material}

\section{Initialization of the Optimization Variables}

We initialize the optimization variables with simple yet physically meaningful priors. Specifically, the latent clean image is initialized as $J^{(0)} = I$, assuming the observed image provides a reasonable starting point for refinement. The transmission map is initialized as $t^{(0)} = 1$, corresponding to a cloud-free assumption that avoids introducing premature attenuation at the early stage of optimization. The atmospheric light is initialized as the spatial mean of the input image, which provides a stable global estimate consistent with its scene-level property. These initialization strategies offer a neutral and well-conditioned starting point, facilitating stable convergence during the unfolding iterations.

\section{Details of Structure-Guided Spectral Cross-Attention Module}

\begin{figure*}[!h]
	\centering
	\includegraphics[width=0.75\linewidth]{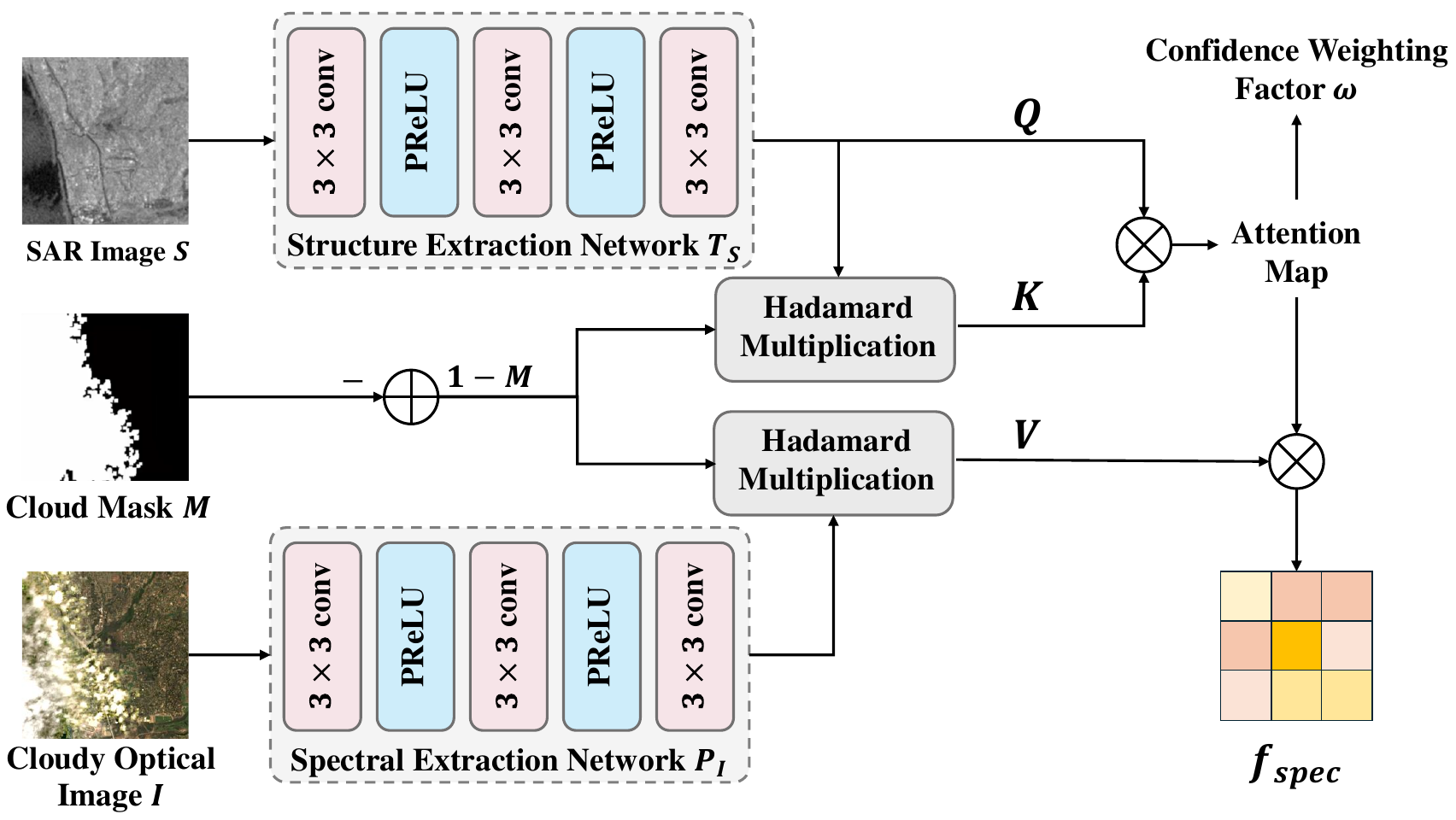}\\
	\caption{Details of Structure-Guided Spectral Cross-Attention Module.}
	\label{SSCM}
\end{figure*}

The network architecture of the Structure-Guided Spectral Cross-Attention Module (SSCM) in Fig.~\ref{model} is shown in Fig.~\ref{SSCM}. The module first extracts structural features from the SAR image via the structure extraction network $T_S$ and evaluates the structural similarity between cloud-covered and cloud-free regions. Based on this similarity, structurally related regions are identified, and then corresponding spectral constraints are imposed through the spectral extraction network $P_I$.

\section{Comprehensive Introduction to Datasets}

\textbf{SEN12MS-CR.} SEN12MS-CR dataset~\cite{sen12mscr} contains 169 non-overlapping regions of interest (ROIs) and approximately 122,218 patch triplets. Each triplet consists of a cloud-contaminated Sentinel-2 optical image, its corresponding cloud-free reference, and a Sentinel-1 SAR image. The Sentinel-2 data include 13 multispectral bands, while Sentinel-1 provides dual-polarized VV and VH channels. Following the standard protocol, ROIs are partitioned into $256 \times 256$ patches. We adopt the full multispectral setting, using all 13 Sentinel-2 bands together with the 2-channel SAR input, rather than restricting to RGB. The optical data are clipped to $[0,10000]$ and normalized to $[0,1]$. The SAR channels are clipped to $[-25,0]$ dB and $[-32.5,0]$ dB, respectively, and then normalized to $[0,1]$. To balance performance and computational cost, 20,000 triplets are randomly selected for training, while the official test set of 7,899 triplets is used for evaluation. The cloud mask $M$ is generated based on normalized blue, cirrus, and visible bands using thresholding rules.

\textbf{SMILE-CR.} SMILE-CR dataset~\cite{smile} is a global multi-modal cloud removal benchmark containing 1,400 cloudy/clear Landsat-8 image pairs, split into 1,000 training, 200 validation, and 200 testing samples. It includes cloudy and cloud-free Landsat-8 images, Sentinel-1 SAR, MODIS observations, and cloud masks. In our experiments, we use a cropped version with $256 \times 256$ patches to maintain consistency with SEN12MS-CR. Only cloudy Landsat-8 images, SAR data, and cloud masks are used as inputs to predict cloud-free Landsat-8 images, while MODIS data are excluded. The resulting dataset contains 3,999 training patches and 800 testing.

\section{More Qualitative Results on \textbf{SEN12MSCR} and \textbf{SMILE-CR} datasets}
\begin{figure*}[!h]
	\centering
	\includegraphics[width=1\linewidth]{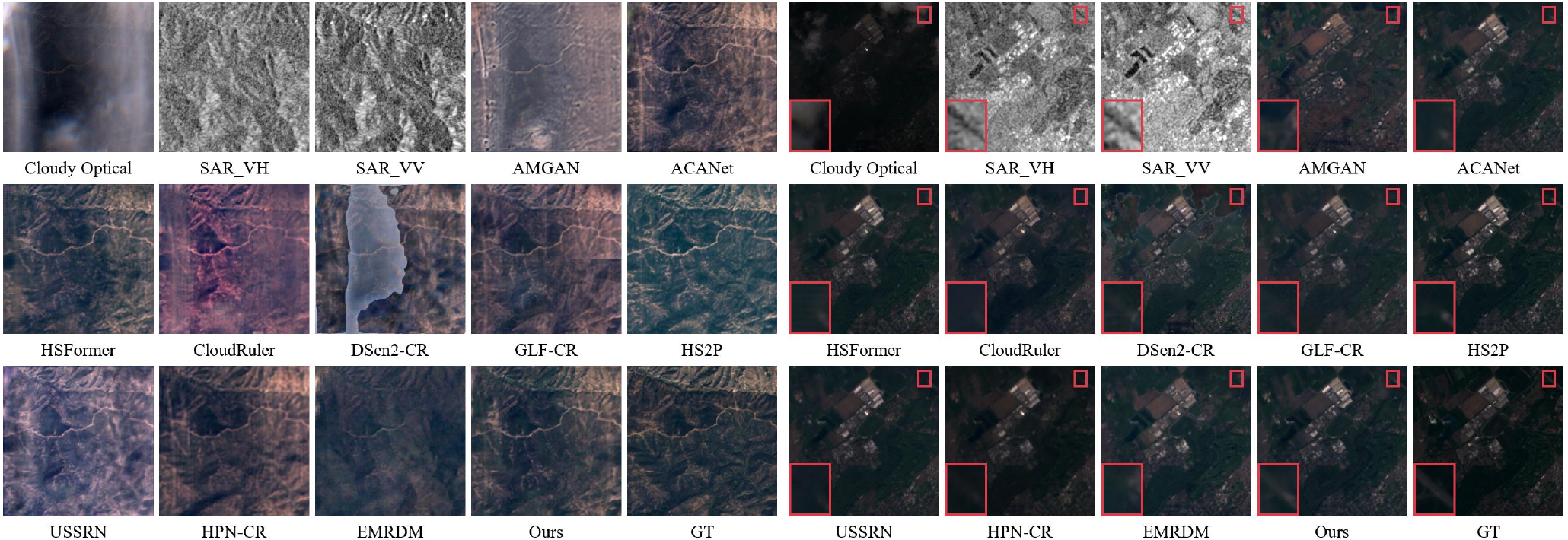}\\
	\vspace{-0.05in}
	\caption{More qualitative comparison of different methods on \textbf{SEN12MS-CR} dataset.}
	\label{supple_sen12mscr}
\end{figure*}

\begin{figure*}[!h]
	\centering
	\includegraphics[width=1\linewidth]{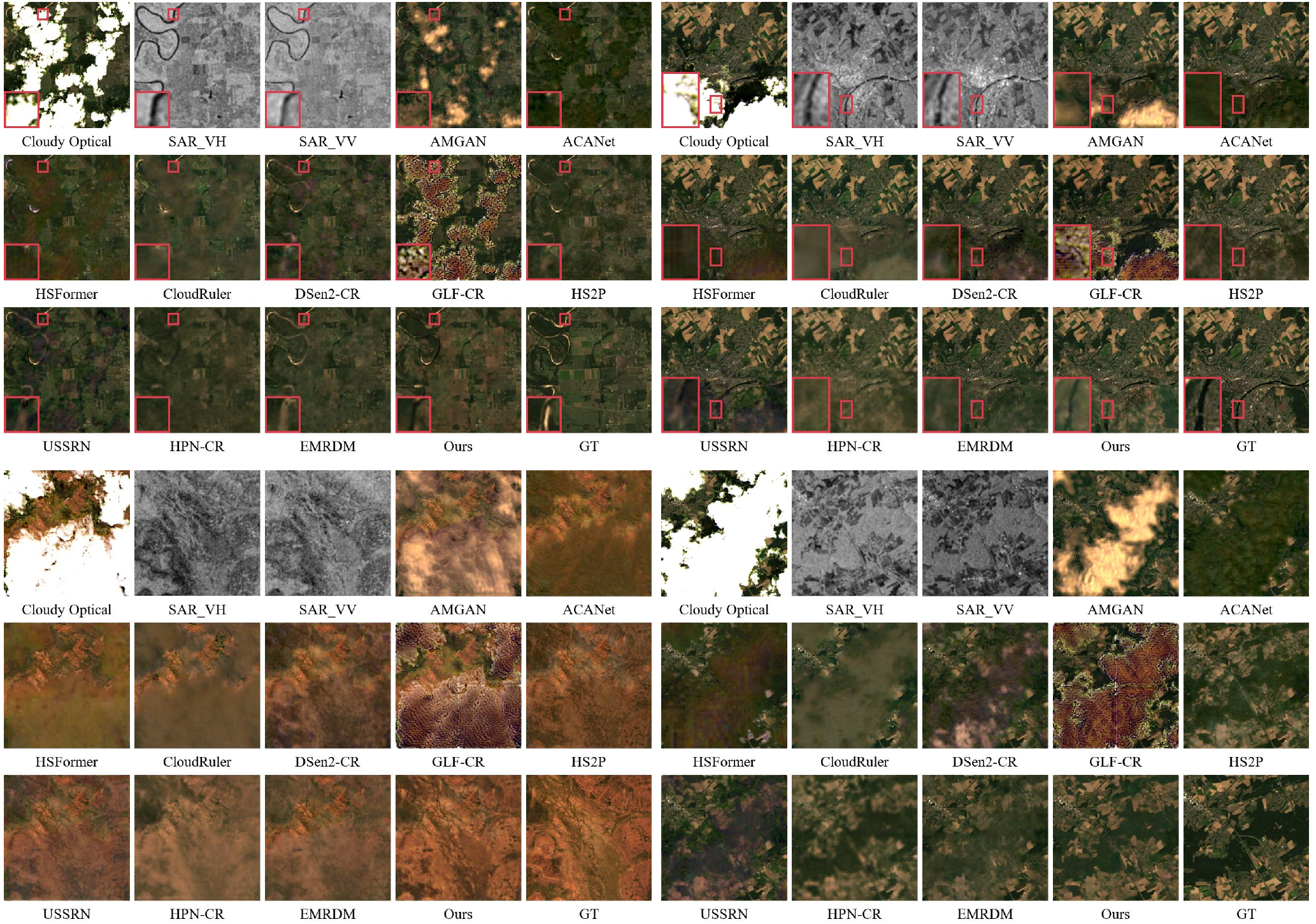}\\
	\vspace{-0.05in}
	\caption{More qualitative comparison of different methods on \textbf{SMILE-CR} dataset.}
	\label{supple_smile}
\end{figure*}

Figs.~\ref{supple_sen12mscr}-\ref{supple_smile} provide additional qualitative comparisons on the two datasets. It can be observed that the proposed method better preserves structural details, with clearer boundaries and more coherent textures than competing approaches. In addition, the reconstructed spectral responses exhibit higher consistency with the ground truth, indicating improved fidelity in both spatial and spectral domains.

\section{Convergence Analysis}

\begin{figure*}[!h]
    \centering
    \begin{adjustbox}{valign=m}
        \begin{minipage}{0.48\linewidth}
            \centering
            \resizebox{\linewidth}{!}{
            \renewcommand\arraystretch{1.15}
            \begin{tabular}{l|ccc} 
                \toprule
                \textbf{Variables} & $J$ & $W$ & $V$ \\
                \midrule
                Iteration Difference ($k_0-k_1$) & 0.249 & 0.323 & 0.338 \\
                Iteration Difference ($k_1-k_2$) & 0.129 & 0.223 & 0.218 \\
                Iteration Difference ($k_2-k_3$) & 0.118 & 0.106 & 0.093 \\
                \bottomrule
            \end{tabular}}
        \end{minipage}
    \end{adjustbox}
    \hfill
    \begin{adjustbox}{valign=m}
        \begin{minipage}{0.42\linewidth}
            \centering
            \includegraphics[width=\linewidth]{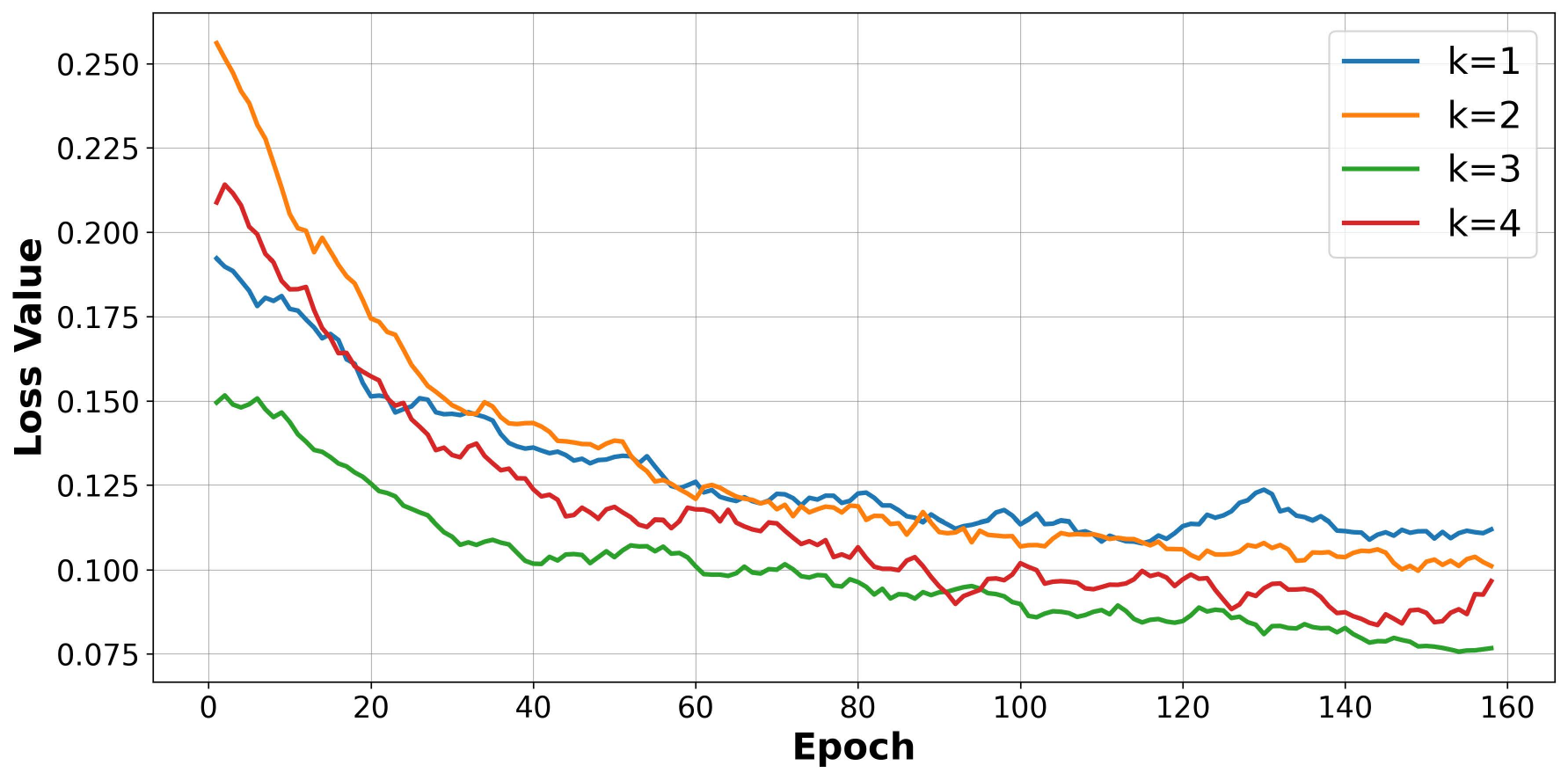}
        \end{minipage}
    \end{adjustbox}

    \vspace{2mm} 
    \begin{minipage}[t]{0.52\linewidth}
        \centering{\footnotesize a) Convergence analysis on different optimization variables.}
    \end{minipage}
    \hfill
    \begin{minipage}[t]{0.45\linewidth}
        \centering{\footnotesize b) Convergence analysis on different iteration numbers.}
    \end{minipage}
    \caption{Convergence analysis of variables and loss. $k_0$ denotes the initial value.}
    \label{convergency_analysis}
\end{figure*}

We conduct a convergence analysis on the training loss of different $K$ and the variations of different optimization variables across iterations, as shown in Fig.~\ref{convergency_analysis}. As illustrated in Fig.~\ref{convergency_analysis}(a), the differences between iterations of the optimization variables gradually decrease as the number of iterations increases, demonstrating the convergence trend of the proposed optimization process. Fig.~\ref{convergency_analysis}(b) further demonstrates that the optimal loss convergence is achieved at $K$=3, in terms of both convergence speed and loss magnitude.

\section{Limitations}

Although existing datasets provide a certain degree of overall scene diversity, the distribution of cloud conditions and land-cover types remains imbalanced. For instance, samples from extreme land-surface scenarios such as glaciers and volcanoes are relatively scarce, which limits the model’s generalization performance in these challenging edge cases. Moreover, although SAR and optical image pairs are geometrically aligned by the datasets in the paper, cross-sensor misalignment may occur in real-world applications and affect cloud removal performance. Thus, incorporating learnable alignment modules or deformable cross-modal registration mechanisms is also an important direction for future work.


\newpage
\section*{NeurIPS Paper Checklist}
\begin{enumerate}
\item {\bf Claims}
    \item[] Question: Do the main claims made in the abstract and introduction accurately reflect the paper's contributions and scope?
    \item[] Answer: \answerYes{}
    \item[] Justification: The abstract and introduction of the paper clearly state the claims made, including the contributions made in the paper and important assumptions and limitations.
    \item[] Guidelines:
    \begin{itemize}
        \item The answer NA means that the abstract and introduction do not include the claims made in the paper.
        \item The abstract and/or introduction should clearly state the claims made, including the contributions made in the paper and important assumptions and limitations. A No or NA answer to this question will not be perceived well by the reviewers.
        \item The claims made should match theoretical and experimental results, and reflect how much the results can be expected to generalize to other settings.
        \item It is fine to include aspirational goals as motivation as long as it is clear that these goals are not attained by the paper.
    \end{itemize}

\item {\bf Limitations}
    \item[] Question: Does the paper discuss the limitations of the work performed by the authors?
    \item[] Answer: \answerYes{}
    \item[] Justification: The paper discusses the limitations of the work performed by the authors, providing a balanced view of the study’s scope and potential areas for improvement.
    \item[] Guidelines:
    \begin{itemize}
        \item The answer NA means that the paper has no limitation while the answer No means that the paper has limitations, but those are not discussed in the paper.
        \item The authors are encouraged to create a separate "Limitations" section in their paper.
        \item The paper should point out any strong assumptions and how robust the results are to violations of these assumptions (e.g., independence assumptions, noiseless settings, model well-specification, asymptotic approximations only holding locally). The authors should reflect on how these assumptions might be violated in practice and what the implications would be.
        \item The authors should reflect on the scope of the claims made, e.g., if the approach was only tested on a few datasets or with a few runs. In general, empirical results often depend on implicit assumptions, which should be articulated.
        \item The authors should reflect on the factors that influence the performance of the approach. For example, a facial recognition algorithm may perform poorly when image resolution is low or images are taken in low lighting. Or a speech-to-text system might not be used reliably to provide closed captions for online lectures because it fails to handle technical jargon.
        \item The authors should discuss the computational efficiency of the proposed algorithms and how they scale with dataset size.
        \item If applicable, the authors should discuss possible limitations of their approach to address problems of privacy and fairness.
        \item While the authors might fear that complete honesty about limitations might be used by reviewers as grounds for rejection, a worse outcome might be that reviewers discover limitations that aren't acknowledged in the paper. The authors should use their best judgment and recognize that individual actions in favor of transparency play an important role in developing norms that preserve the integrity of the community. Reviewers will be specifically instructed to not penalize honesty concerning limitations.
    \end{itemize}

\item {\bf Theory assumptions and proofs}
    \item[] Question: For each theoretical result, does the paper provide the full set of assumptions and a complete (and correct) proof?
    \item[] Answer: \answerYes{}
	\item[] Justification: All the theorems, formulas, and proofs in the paper have been numbered and cross-referenced, and all assumptions have been clearly stated or referenced in the statement of any theorems.
    \item[] Guidelines:
    \begin{itemize}
        \item The answer NA means that the paper does not include theoretical results.
        \item All the theorems, formulas, and proofs in the paper should be numbered and cross-referenced.
        \item All assumptions should be clearly stated or referenced in the statement of any theorems.
        \item The proofs can either appear in the main paper or the supplemental material, but if they appear in the supplemental material, the authors are encouraged to provide a short proof sketch to provide intuition.
        \item Inversely, any informal proof provided in the core of the paper should be complemented by formal proofs provided in appendix or supplemental material.
        \item Theorems and Lemmas that the proof relies upon should be properly referenced.
    \end{itemize}

    \item {\bf Experimental result reproducibility}
    \item[] Question: Does the paper fully disclose all the information needed to reproduce the main experimental results of the paper to the extent that it affects the main claims and/or conclusions of the paper (regardless of whether the code and data are provided or not)?
    \item[] Answer: \answerYes{}
	\item[] Justification: The paper has fully disclosed all the information needed to reproduce the main experimental results of the paper to the extent that it affects the main claims and conclusions of the paper.
    \item[] Guidelines:
    \begin{itemize}
        \item The answer NA means that the paper does not include experiments.
        \item If the paper includes experiments, a No answer to this question will not be perceived well by the reviewers: Making the paper reproducible is important, regardless of whether the code and data are provided or not.
        \item If the contribution is a dataset and/or model, the authors should describe the steps taken to make their results reproducible or verifiable.
        \item Depending on the contribution, reproducibility can be accomplished in various ways. For example, if the contribution is a novel architecture, describing the architecture fully might suffice, or if the contribution is a specific model and empirical evaluation, it may be necessary to either make it possible for others to replicate the model with the same dataset, or provide access to the model. In general. releasing code and data is often one good way to accomplish this, but reproducibility can also be provided via detailed instructions for how to replicate the results, access to a hosted model (e.g., in the case of a large language model), releasing of a model checkpoint, or other means that are appropriate to the research performed.
        \item While NeurIPS does not require releasing code, the conference does require all submissions to provide some reasonable avenue for reproducibility, which may depend on the nature of the contribution. For example
        \begin{enumerate}
            \item If the contribution is primarily a new algorithm, the paper should make it clear how to reproduce that algorithm.
            \item If the contribution is primarily a new model architecture, the paper should describe the architecture clearly and fully.
            \item If the contribution is a new model (e.g., a large language model), then there should either be a way to access this model for reproducing the results or a way to reproduce the model (e.g., with an open-source dataset or instructions for how to construct the dataset).
            \item We recognize that reproducibility may be tricky in some cases, in which case authors are welcome to describe the particular way they provide for reproducibility. In the case of closed-source models, it may be that access to the model is limited in some way (e.g., to registered users), but it should be possible for other researchers to have some path to reproducing or verifying the results.
        \end{enumerate}
    \end{itemize}

\item {\bf Open access to data and code}
    \item[] Question: Does the paper provide open access to the data and code, with sufficient instructions to faithfully reproduce the main experimental results, as described in supplemental material?
    \item[] Answer: \answerYes{}
    \item[] Justification: We will open access to the code after the paper is accepted.
    \item[] Guidelines:
    \begin{itemize}
        \item The answer NA means that paper does not include experiments requiring code.
        \item Please see the NeurIPS code and data submission guidelines (\url{https://nips.cc/public/guides/CodeSubmissionPolicy}) for more details.
        \item While we encourage the release of code and data, we understand that this might not be possible, so “No” is an acceptable answer. Papers cannot be rejected simply for not including code, unless this is central to the contribution (e.g., for a new open-source benchmark).
        \item The instructions should contain the exact command and environment needed to run to reproduce the results. See the NeurIPS code and data submission guidelines (\url{https://nips.cc/public/guides/CodeSubmissionPolicy}) for more details.
        \item The authors should provide instructions on data access and preparation, including how to access the raw data, preprocessed data, intermediate data, and generated data, etc.
        \item The authors should provide scripts to reproduce all experimental results for the new proposed method and baselines. If only a subset of experiments are reproducible, they should state which ones are omitted from the script and why.
        \item At submission time, to preserve anonymity, the authors should release anonymized versions (if applicable).
        \item Providing as much information as possible in supplemental material (appended to the paper) is recommended, but including URLs to data and code is permitted.
    \end{itemize}

\item {\bf Experimental setting/details}
    \item[] Question: Does the paper specify all the training and test details (e.g., data splits, hyperparameters, how they were chosen, type of optimizer, etc.) necessary to understand the results?
    \item[] Answer: \answerYes{}
	\item[] Justification: We have specified all the training and test details necessary to understand the results.
    \item[] Guidelines:
    \begin{itemize}
        \item The answer NA means that the paper does not include experiments.
        \item The experimental setting should be presented in the core of the paper to a level of detail that is necessary to appreciate the results and make sense of them.
        \item The full details can be provided either with the code, in appendix, or as supplemental material.
    \end{itemize}

\item {\bf Experiment statistical significance}
    \item[] Question: Does the paper report error bars suitably and correctly defined or other appropriate information about the statistical significance of the experiments?
    \item[] Answer: \answerYes{}
	\item[] Justification: The paper reports error bars and other appropriate information about the statistical significance of the experiments, ensuring the reliability and validity of the results.
    \item[] Guidelines:
    \begin{itemize}
        \item The answer NA means that the paper does not include experiments.
        \item The authors should answer "Yes" if the results are accompanied by error bars, confidence intervals, or statistical significance tests, at least for the experiments that support the main claims of the paper.
        \item The factors of variability that the error bars are capturing should be clearly stated (for example, train/test split, initialization, random drawing of some parameter, or overall run with given experimental conditions).
        \item The method for calculating the error bars should be explained (closed form formula, call to a library function, bootstrap, etc.)
        \item The assumptions made should be given (e.g., Normally distributed errors).
        \item It should be clear whether the error bar is the standard deviation or the standard error of the mean.
        \item It is OK to report 1-sigma error bars, but one should state it. The authors should preferably report a 2-sigma error bar than state that they have a 96\% CI, if the hypothesis of Normality of errors is not verified.
        \item For asymmetric distributions, the authors should be careful not to show in tables or figures symmetric error bars that would yield results that are out of range (e.g. negative error rates).
        \item If error bars are reported in tables or plots, The authors should explain in the text how they were calculated and reference the corresponding figures or tables in the text.
    \end{itemize}

\item {\bf Experiments compute resources}
    \item[] Question: For each experiment, does the paper provide sufficient information on the computer resources (type of compute workers, memory, time of execution) needed to reproduce the experiments?
    \item[] Answer: \answerYes{}
	\item[] Justification: The paper provides sufficient information on the computer resource for experiments.
    \item[] Guidelines:
    \begin{itemize}
        \item The answer NA means that the paper does not include experiments.
        \item The paper should indicate the type of compute workers CPU or GPU, internal cluster, or cloud provider, including relevant memory and storage.
        \item The paper should provide the amount of compute required for each of the individual experimental runs as well as estimate the total compute.
        \item The paper should disclose whether the full research project required more compute than the experiments reported in the paper (e.g., preliminary or failed experiments that didn't make it into the paper).
    \end{itemize}

\item {\bf Code of ethics}
    \item[] Question: Does the research conducted in the paper conform, in every respect, with the NeurIPS Code of Ethics \url{https://neurips.cc/public/EthicsGuidelines}?
    \item[] Answer: \answerYes{}
	\item[] Justification: The research conducted in the paper conforms in every respect with the NeurIPS Code of Ethics, ensuring ethical standards are upheld throughout the study.
    \item[] Guidelines:
    \begin{itemize}
        \item The answer NA means that the authors have not reviewed the NeurIPS Code of Ethics.
        \item If the authors answer No, they should explain the special circumstances that require a deviation from the Code of Ethics.
        \item The authors should make sure to preserve anonymity (e.g., if there is a special consideration due to laws or regulations in their jurisdiction).
    \end{itemize}

\item {\bf Broader impacts}
    \item[] Question: Does the paper discuss both potential positive societal impacts and negative societal impacts of the work performed?
    \item[] Answer: \answerYes{}
	\item[] Justification: The paper discusses both potential positive societal impacts and negative societal impacts of the work performed, providing a comprehensive evaluation of the study’s broader implications.
    \item[] Guidelines:
    \begin{itemize}
        \item The answer NA means that there is no societal impact of the work performed.
        \item If the authors answer NA or No, they should explain why their work has no societal impact or why the paper does not address societal impact.
        \item Examples of negative societal impacts include potential malicious or unintended uses (e.g., disinformation, generating fake profiles, surveillance), fairness considerations (e.g., deployment of technologies that could make decisions that unfairly impact specific groups), privacy considerations, and security considerations.
        \item The conference expects that many papers will be foundational research and not tied to particular applications, let alone deployments. However, if there is a direct path to any negative applications, the authors should point it out. For example, it is legitimate to point out that an improvement in the quality of generative models could be used to generate deepfakes for disinformation. On the other hand, it is not needed to point out that a generic algorithm for optimizing neural networks could enable people to train models that generate Deepfakes faster.
        \item The authors should consider possible harms that could arise when the technology is being used as intended and functioning correctly, harms that could arise when the technology is being used as intended but gives incorrect results, and harms following from (intentional or unintentional) misuse of the technology.
        \item If there are negative societal impacts, the authors could also discuss possible mitigation strategies (e.g., gated release of models, providing defenses in addition to attacks, mechanisms for monitoring misuse, mechanisms to monitor how a system learns from feedback over time, improving the efficiency and accessibility of ML).
    \end{itemize}

\item {\bf Safeguards}
    \item[] Question: Does the paper describe safeguards that have been put in place for responsible release of data or models that have a high risk for misuse (e.g., pretrained language models, image generators, or scraped datasets)?
    \item[] Answer: \answerYes{}
	\item[] Justification: The paper describes safeguards that have been put in place for the responsible release of data or models that have a high risk for misuse, ensuring that appropriate measures are taken to mitigate potential risks.
    \item[] Guidelines:
    \begin{itemize}
        \item The answer NA means that the paper poses no such risks.
        \item Released models that have a high risk for misuse or dual-use should be released with necessary safeguards to allow for controlled use of the model, for example by requiring that users adhere to usage guidelines or restrictions to access the model or implementing safety filters.
        \item Datasets that have been scraped from the Internet could pose safety risks. The authors should describe how they avoided releasing unsafe images.
        \item We recognize that providing effective safeguards is challenging, and many papers do not require this, but we encourage authors to take this into account and make a best faith effort.
    \end{itemize}

\item {\bf Licenses for existing assets}
    \item[] Question: Are the creators or original owners of assets (e.g., code, data, models), used in the paper, properly credited and are the license and terms of use explicitly mentioned and properly respected?
    \item[] Answer: \answerYes{}
	\item[] Justification: The creators or original owners of assets used in the paper are properly credited, and the license and terms of use are explicitly mentioned and properly respected.
    \item[] Guidelines:
    \begin{itemize}
        \item The answer NA means that the paper does not use existing assets.
        \item The authors should cite the original paper that produced the code package or dataset.
        \item The authors should state which version of the asset is used and, if possible, include a URL.
        \item The name of the license (e.g., CC-BY 4.0) should be included for each asset.
        \item For scraped data from a particular source (e.g., website), the copyright and terms of service of that source should be provided.
        \item If assets are released, the license, copyright information, and terms of use in the package should be provided. For popular datasets, \url{paperswithcode.com/datasets} has curated licenses for some datasets. Their licensing guide can help determine the license of a dataset.
        \item For existing datasets that are re-packaged, both the original license and the license of the derived asset (if it has changed) should be provided.
        \item If this information is not available online, the authors are encouraged to reach out to the asset's creators.
    \end{itemize}

\item {\bf New assets}
    \item[] Question: Are new assets introduced in the paper well documented and is the documentation provided alongside the assets?
    \item[] Answer:\answerYes{}
	\item[] Justification: The new assets introduced in the paper are well documented, and the documentation is provided alongside the assets.
    \item[] Guidelines:
    \begin{itemize}
        \item The answer NA means that the paper does not release new assets.
        \item Researchers should communicate the details of the dataset/code/model as part of their submissions via structured templates. This includes details about training, license, limitations, etc.
        \item The paper should discuss whether and how consent was obtained from people whose asset is used.
        \item At submission time, remember to anonymize your assets (if applicable). You can either create an anonymized URL or include an anonymized zip file.
    \end{itemize}

\item {\bf Crowdsourcing and research with human subjects}
    \item[] Question: For crowdsourcing experiments and research with human subjects, does the paper include the full text of instructions given to participants and screenshots, if applicable, as well as details about compensation (if any)?
    \item[] Answer: \answerNA{}
	\item[] Justification: The paper does not involve crowdsourcing nor research with human subjects.
    \item[] Guidelines:
    \begin{itemize}
        \item The answer NA means that the paper does not involve crowdsourcing nor research with human subjects.
        \item Including this information in the supplemental material is fine, but if the main contribution of the paper involves human subjects, then as much detail as possible should be included in the main paper.
        \item According to the NeurIPS Code of Ethics, workers involved in data collection, curation, or other labor should be paid at least the minimum wage in the country of the data collector.
    \end{itemize}

\item {\bf Institutional review board (IRB) approvals or equivalent for research with human subjects}
    \item[] Question: Does the paper describe potential risks incurred by study participants, whether such risks were disclosed to the subjects, and whether Institutional Review Board (IRB) approvals (or an equivalent approval/review based on the requirements of your country or institution) were obtained?
    \item[] Answer: \answerNA{}
	\item[] Justification: The paper does not involve crowdsourcing nor research with human subjects.
    \item[] Guidelines:
    \begin{itemize}
        \item The answer NA means that the paper does not involve crowdsourcing nor research with human subjects.
        \item Depending on the country in which research is conducted, IRB approval (or equivalent) may be required for any human subjects research. If you obtained IRB approval, you should clearly state this in the paper.
        \item We recognize that the procedures for this may vary significantly between institutions and locations, and we expect authors to adhere to the NeurIPS Code of Ethics and the guidelines for their institution.
        \item For initial submissions, do not include any information that would break anonymity (if applicable), such as the institution conducting the review.
    \end{itemize}

\item {\bf Declaration of LLM usage}
    \item[] Question: Does the paper describe the usage of LLMs if it is an important, original, or non-standard component of the core methods in this research? Note that if the LLM is used only for writing, editing, or formatting purposes and does not impact the core methodology, scientific rigorousness, or originality of the research, declaration is not required.
    \item[] Answer: \answerNA{}
	\item[] Justification: The core method development in the paper does not involve LLMs as any important, original, or non-standard components.
    \item[] Guidelines:
    \begin{itemize}
        \item The answer NA means that the core method development in this research does not involve LLMs as any important, original, or non-standard components.
        \item Please refer to our LLM policy (\url{https://neurips.cc/Conferences/2025/LLM}) for what should or should not be described.
    \end{itemize}

\end{enumerate}

\end{document}